\documentclass[10pt,twocolumn,letterpaper]{article}

\usepackage{wacv}
\usepackage{times}
\usepackage{epsfig}
\usepackage{graphicx}
\usepackage{amsmath}
\usepackage{amssymb}
\usepackage{multirow}
\usepackage{breqn}
\usepackage{xcolor}

\def\wacvPaperID{****} 

\wacvfinalcopy 

\ifwacvfinal
\def\assignedStartPage{1} 
\fi

\ifwacvfinal
\usepackage[breaklinks=true,bookmarks=false]{hyperref}
\else
\usepackage[pagebackref=true,breaklinks=true,colorlinks,bookmarks=false]{hyperref}
\fi

\ifwacvfinal
\else
\fi

\begin{document}

\title{Comprehensive Online Network Pruning via Learnable Scaling Factors}

\author{Muhammad Umair Haider\\
Lahore University of Management Sciences\\
Lahore Cantt, Pakistan\\
{\tt\small 18030031@lums.edu.pk}
\and
Murtaza Taj\\
Lahore University of Management Sciences\\
Lahore Cantt, Pakistan\\
{\tt\small murtaza.taj@lums.edu.pk}
}

\maketitle

\begin{abstract}
One of the major challenges in deploying deep neural network architectures is their size which has an adverse effect on their inference time and memory requirements. Deep CNNs can either be pruned width-wise by removing filters based on their importance or depth-wise by removing layers and blocks. Width wise pruning (filter pruning) is commonly performed via learnable gates or switches and sparsity regularizers whereas pruning of layers has so far been performed arbitrarily by manually designing a smaller network usually referred to as a student network. We propose a comprehensive pruning strategy that can perform both width-wise as well as depth-wise pruning. This is achieved by introducing gates at different granularities (neuron, filter, layer, block) which are then controlled via an objective function that simultaneously performs pruning at different granularity during each forward pass. Our approach is applicable to wide-variety of architectures without any constraints on spatial dimensions or connection type (sequential, residual, parallel or inception). Our method has resulted in a compression ratio of $70\%$ to $90\%$ without noticeable loss in accuracy when evaluated on benchmark datasets.
\end{abstract}

\section{Introduction}
With the increasing complexity of problems the size of deep convolutional neural networks (CNNs) is also increasing exponentially to a point where CNNs are so huge that they are already hitting the computational limits of the devices on which they are running. There is a significant amount of redundancy in these huge CNNs that can be removed (i.e: pruned) to accelerate them. It is well known that neural nets are nature inspired, it is also a fact that pruning these neural nets is also nature inspired. According to human brain development studies~\cite{1} “\textit{one of the fundamental phenomena in brain development is the reduction of the amount of synapses that occurs between early childhood and puberty}” and is commonly referred to as \textit{synaptic pruning}. One of the earliest work on pruning neural networks was performed by LeCun in the 1990s~\cite{2} in which network connections were pruned based on \textbf{$2^{nd}$ derivative} of weights. {Development of CNNs in 1998 again by LeCun~\cite{LeCun1998} and their successful application in ImageNet classification challenge in 2012}~\cite{alexnet} resulted in a revival of this concept of pruning in 2015~\cite{3}. They used \textbf{L1 norm} to prune away lighter weights.

\begin{figure}
    \centering
    \begin{tabular}{cc}
        \includegraphics[width=.22\columnwidth]{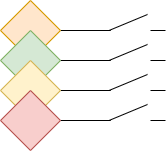} &  
        \includegraphics[width=.45\columnwidth]{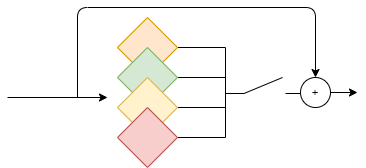} \\
        (a) Filter Pruning~\cite{27} & (b) Layer Pruning \\
        \includegraphics[width=.5\columnwidth]{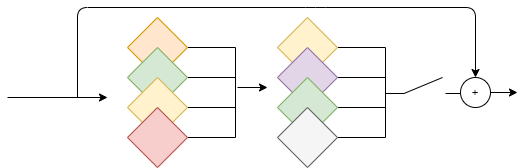} &  
        \includegraphics[width=.5\columnwidth]{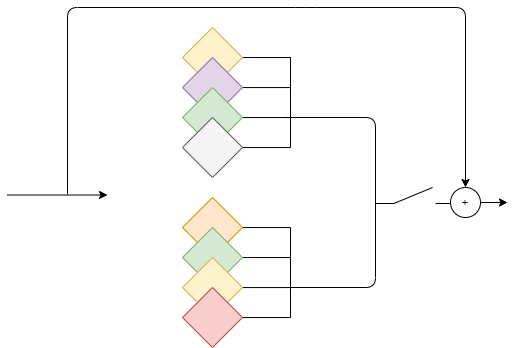} \\
        (c) Branch Pruning~\cite{resnet} & (d) Block Pruning \\
        \multicolumn{2}{c}{\includegraphics[width=1.0\columnwidth]{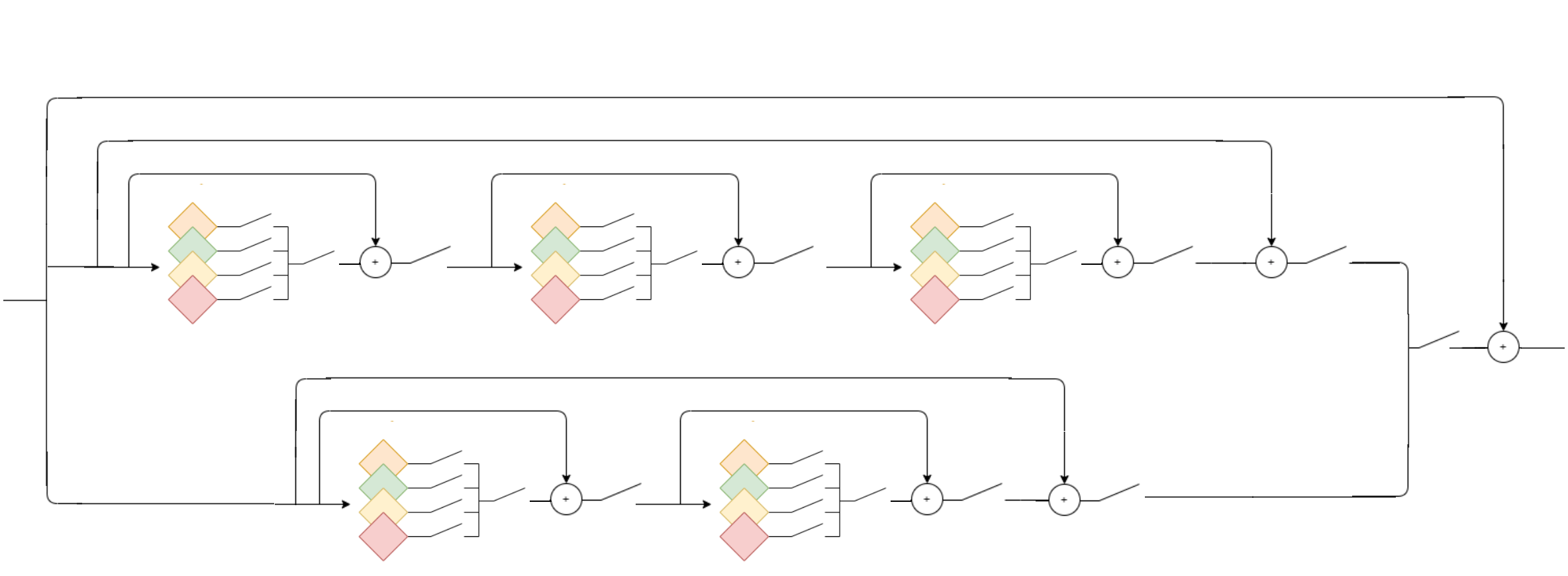}}\\
        \multicolumn{2}{c}{(e) Comprehensive Pruning.}
    \end{tabular}
    \caption{Sample illustration showing pruning at different granularities. (a) Filter pruning. (b) Layer pruning. (c) Branch pruning. (d) Block pruning. (e) Comprehensive Pruning.}
    \label{fig:my_label}
    \vspace{-.5cm}
\end{figure}

Both of these pruning techniques~\cite{2,3} pruned individual neurons from the network except that in the recent method~\cite{3} \textbf{L1 norm} was used to prune away lighter weights. Although these methods pruned away a large number of weights from the network but they had a major shortcoming of introducing sparsity. More recent literature addressed this concern first by pruning filters as convolution operations constitutes the main computational burden of a CNN and second by using sparsity regularizers~\cite{4,5,6,7,8,9,17,18,22,27,GateNIPS2019_8486}.  

There are two major methodologies in neural net compression i) structured pruning and knowledge distillation. The existing work on structured pruning~\cite{4,5,6,7,8,9,17,18,22,27,GateNIPS2019_8486} addresses only the width of the layer by removing filters based on their importance. However, selecting what to prune from large CNNs is an NP-Hard problem, to find the optimal solution one would need to rank each filter by turning it off and perform inference using all the samples. Thus, if we are to prune network structures like filters we would need to try out all possibilities of filter configurations and choose the best one that has the least effect on the loss. This problem is addressed in literature by a sparsity loss and learnable dropout parameters.

In knowledge distillation a large cumbersome CNN called teacher network is used to train a smaller network referred to as student network~\cite{HintonPaper}. The student network is designed by manually removing structured elements such as filters, layers, blocks from an existing network in an adhoc manner. Despite several successes in training a student network~\cite{wang2018adversarial, hinton2015distilling, you2017learning, you2017learning}, designing a student network is still an open problem and is usually performed via hit and try which is a tedious process.

In our work we overcome these challenges by proposing a comprehensive pruning strategy that performs both width-wise (filters) as well as depth-wise (layers, branch, blocks) pruning in an online manner. For width-wise pruning we utilize the idea of filter pruning~\cite{27,GateNIPS2019_8486} whereas for depth-wise pruning we proposed layer, branch and block pruning schemes. The main contributions of our work are as follows:

\begin{itemize}
    \item We extend the width-wise pruning strategy via learnable dropout parameters and propose a depth-wise pruning strategy.
    \item Ours is a comprehensive pruning strategy that performs both width-wise and depth-wise pruning by introducing gates with learnable scaling factor at different granularities (filter, layer, sequence of layers or branch, multi-branched or blocks).
    \item Instead of pruning layers based on the remaining number of filters, ours is a global pruning strategy that eliminates redundant layers considering the overall structure of the network. 
    \item Our method can also be used to eliminate entire blocks such as inception blocks from the network and is applicable to wide-variety of architectures.
    \item We also propose an objective function that simultaneously eliminates filters, layers and blocks by introducing sparsity in gates with learnable scaling factor. 
\end{itemize}

\section{Related Work}

\subsection{Offline Pruning}
Numerous offline filter pruning techniques to rank and prune filters have been proposed. These methods can be categorized into two groups: those that analyze the filters themselves and those that perform pruning based on filter activations. Ranking and pruning based on filter analysis is performed using various measures including {L1 norm}~\cite{4}, {L1-norm and standard deviation}~\cite{5}, Entropy~\cite{6}, Geometric Median~\cite{7} and average percentage of zero activations~\cite{8}. A new optimization method that enforces correlation among filters and then safely removes the redundant filters has also been proposed~\cite{9}. Filter activation maps inform us about whether the certain filter fires on a given dataset. Thus a set of filters which may have a very distinctive shape may result in a very low activation and thus can be pruned for a particular dataset. Similarly, similar filters will result in similar activations and thus redundancy can also be identified via activation maps.

Low-rank feature maps~\cite{17} and LASSO regression based channel selection~\cite{18} have also been proposed to perform pruning through activation maps. Similarly, by forcing the outputs of some channels to be constant allow us to prune those constant channels~\cite{22}. Additional losses have also been proposed to increase the discriminative power of intermediate layers and then select the most discriminative channels for each layer while rejecting the other channels~\cite{24}. The major drawback of offline pruning strategies is that in offline structured pruning the network convergence is stopped to generate a new pruned network and then this pruned network is converged again and pruned again, this periodic pruning process may result in a pruned network that fails to achieve the same accuracy as compared to the original network. These issues are addressed via online pruning which is discussed next.

\subsection{Online Pruning via Scaling Factor}
Structured pruning via scaling factors is a technique where the output of a structure is multiplied to a scaling factor. In online structured pruning, the pruning is modelled as a task of learning this scaling factor associated to each filter~\cite{27,GateNIPS2019_8486}. This scaling factor acts as an on/off switch which in its off state mimics the elimination of the filter. Regularization is usually applied to these scaling factors to prune the corresponding structures from the network~\cite{27,GateNIPS2019_8486}. Several schemes have been proposed to update the values of these scaling factors during training. For e.g. differentiable Markov Chains has been used to find prunable filters where each node of the chain is the state of a filter either on or off~\cite{21}. Similarly, if a noise across layers is represented as a Markov Chain, then its posterior can be used to reflect inter-layer dependency which act as a measure for pruning\cite{26}. Rather than pruning the network to the maximum possible limit without noticeable loss in accuracy, pruning can also be performed to achieve a certain budget. In such cases knowledge distillation loss~\cite{HintonPaper} with budget aware regularizers have been shown to perform the desired amount of compression~\cite{27}.

Most of the online and offline pruning strategies are limited to width-wise or filter pruning only. There has been multiple attempts of reducing the depth of the network. This is usually performed by removing several layers and filters from the network in an adhoc manners followed by retraining (for e.g. Yolo vs. Tiny Yolo~\cite{yolo}\cite{tinyyolo}, ResNet-101 vs. ResNet-56~\cite{resnet}).  This adversely affect the representative power of the network and creates challenges during retraining. In more recent literature the issue of retraining is addressed using knowledge distillation~\cite{HintonPaper} where the unpruned network called teacher network is used to train a smaller student network via a loss function that minimizes both the data loss as well as the discrepancy between soft logits of teacher and student network. In this work, instead of simplifying a network in an adhoc manner, we extend the idea of online pruning via scaling factors and show how these can be applied to any structured element of any granularity.


\section{Methodology}
In this work we extend the idea of pruning via scaling factors and show that it can be applied to any structured element of any granularity level. It can be applied to coarse granular structures like layers, blocks, branches or even sub-networks to fine granular structures like filters or even individual neurons. 

Although in our work we used sparsity regularizer, one can even use different regularizations for different granularity levels simultaneously (e.g: applying one regularization to blocks and another to filters within the block).

\subsection{Filter Pruning} 
The idea of online filters pruning via scaling factors is fundamentally similar to the concept of dropout~\cite{JMLR:v15:srivastava14a} in which units (along with their connections) are randomly dropped from the neural network during training. Considering a neural network with $L$ hidden layers $\mathbf{h}_l$ where $l\in \{1,\cdots,L\}$. The feed-forward operation $f(.)$ of this network, which is typically convolution in case of CNNs, with dropout is represented as:
%
\begin{equation}
\mathbf{h}_l = f(\mathbf{h}_{l-1}\odot \mathbf{z}_{l-1}).
\end{equation}
In case of dropout~\cite{JMLR:v15:srivastava14a} $\mathbf{z}_{l-1}$ is a tensor of independent random variables i.i.d. of a Bernoulli distribution $q(z)$. Literature on filter pruning~\cite{27} redefined $\mathbf{z}_{l-1}$ via (reparameterization trick~\cite{NIPS2015_5666}) using a continuous differentiable function $g$ having learnable parameters $\Phi$ (also known as scaling factors) as:
\begin{equation}
\mathbf{h}_l = f(\mathbf{h}_{l-1}\odot g(\Phi_{l-1}^f,\epsilon))
\end{equation}
where $g$ is stochastic w.r.t a random variable $\epsilon$ typically sampled from $\mathcal{N} (0, 1)$ or $\mathcal{U} (0, 1)$.

In our design of filter pruning we also place \textit{pruning tensor vectors} which belong to the \textit{uniform distribution } in front of every convolution layer which are multiplied with the output of convolution layer to zero out (i.e: prune) unwanted kernels. These pruning tensors are placed as \textit{scaling factors} in the network after convolution layers so that the output of each filter is multiplied by a corresponding scaling factor. Setting the value of these pruning tensors to zero will essentially mean pruning their corresponding structure element from the network in both forward and backward pass. We update the value of network parameters including $Phi$ by simply using sparsity loss $\mathcal{L}_s(\Phi^f)$, thus our loss function is defined as:

\begin{equation}
\mathcal{L}(W,\Phi) = \mathcal{L}_D(W,\Phi) + \lambda^f\mathcal{L}_s(\Phi^f)
\end{equation}
where $\mathcal{L}_D(W,\Phi)$ is a data loss and $\lambda^f$ is the Lagrangian multiplier and the superscript $f$ indicates that these parameters are associated with filter pruning. Although $\mathcal{L}_D$ could be any appropriate loss function however in our experiments we chose it to be cross entropy loss. 

\subsection{Layers Pruning}


Layer pruning has only been scarcely addressed in literature as a special case of filter pruning~\cite{27}. In these methods a layer is eliminated when all the values in $\Phi$ approaches close to zero or all the filters are turned off. Since this blocks further propagation of values across the network, it has only shown to work in case of residual networks~\cite{resnet}. This makes layer pruning only a coincidental event without any explicit incentive towards loss minimization.

We proposed to explicitly model layer pruning by incentivising the network to reduce its depth. We proposed two modifications, first, we developed an algorithm that introduces shortcut connections with each convolution layer in the network and is thus applicable to both ResNets as well as other networks. Secondly we extend the idea of filter pruning for layer pruning by introducing a learnable layer dropout parameter $\phi^l_l$. This pruning parameter is multiplied with the whole convolution layer’s output so the new equations for any output would be: 
\begin{equation}
\mathbf{h}_l = f(\mathbf{h}_{l-1})g(\Phi_{l}^l,\epsilon))+\mathbf{h}_{l-1}
\end{equation}
We then extended the sparse dropout learning for layer pruning by defining a layer pruning loss function as:
\begin{equation}
\mathcal{L}(W,\Phi) = \mathcal{L}_D(W,\Phi) + \lambda^l\mathcal{L}_s(\Phi^l)
\end{equation}
where $\mathcal{L}_s(\Phi^l)$ is the layer sparsity loss and $\lambda^l$ is the Lagrangian multiplier.
This enables the network to decide what the \textit{depth} should be and which layers can be skipped. The combined loss for filter and layer pruning can then be defined as:
\begin{equation}
\mathcal{L}(W,\Phi) = \mathcal{L}_D(W,\Phi) + \lambda^f\mathcal{L}_s(\Phi^f) + \lambda^l\mathcal{L}_s(\Phi^l).
\label{eq:filterlayer}
\end{equation}
%
%


\begin{figure}[t]
    \centering
    \includegraphics[width=.65\columnwidth]{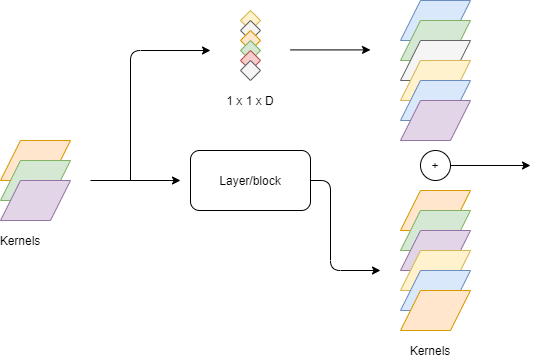}
    \caption{Sample illustration showing convolution of $1\times 1\times d_1$ where $d_1$ is the depth of destination layer.}
    \label{fig:1x1xD}
\end{figure}

\textbf{Handling Dimension Mismatch}: As discussed earlier, in our design we add a shortcut connection across layer to be pruned to pass data in the network, but to do this the dimensions of the output of the previous layer should be the same as the dimensions of the output of the current layer.  This leads to two key challenges in performing layer pruning, i) change in spatial dimension and ii) change in depth dimension. If the depth does not match we can use a convolution of $1\times 1\times d_1$ where $d_1$ is the depth of destination layer (see Fig.~\ref{fig:1x1xD}) If spatial dimension does not match we can use pooling. If both does not match we can use pooling followed by $1\times 1\times d_1$ where $d_1$ to resize the output of the previous layer.
\subsection{Branch Pruning}
The idea of branch pruning was first introduced by ResNets~\cite{resnet} primarily to address the issue of vanishing gradients in deeper networks. The idea being that if neural networks can approximate a complicated function they can also approximate residual function $f(\mathbf{h}_{l-1}) - \mathbf{h}_{l-1}$ which can be defined as:
\begin{equation}
    \mathbf{h}_{l} = f(\mathbf{h}_{l-1}) + \mathbf{h}_{l-1}.    
\end{equation}
This operation is performed by a shortcut connection and element-wise addition. Due to this shortcut connection if identity mappings are optimal, the solvers may simply drive the weights of the multiple nonlinear layers toward zero to approach identity mappings.

Similar to layer pruning, here as well we incentivise the network to reduce its depth by completely eliminating a particular branch. The type of pruning can be beneficial in networks with multiple branches such as within inception block~\cite{inception} or in those networks that use ensembles of multiple parallel branches~\cite{NIPS2018_7980}. Here again we propose two modifications, first, we developed an algorithm that introduces shortcut connections after a branched out sequence of layers if branches are not already present in the network. Secondly we extend the idea of filter pruning and layer pruning for branch pruning by introducing a learnable branch dropout parameter $\phi^r_l$. This pruning parameter is multiplied with the whole branch. Let a branch is defined as $\mathbf{r}_{l-1} = \mathbf{h}_{l-n}:\mathbf{h}_{l-1}$, so the new equations for any output would be: 
\begin{equation}
\mathbf{h}_l = f(\mathbf{r}_{l-1})g(\Phi_{l}^r,\epsilon))+\mathbf{h}_{l-n}.
\end{equation}
We then extended the sparse dropout learning for layer pruning by defining a layer pruning loss function as:
\begin{equation}
\mathcal{L}(W,\Phi) = \mathcal{L}_D(W,\Phi) + \lambda^r\mathcal{L}_s(\Phi^r)
\end{equation}
where $\mathcal{L}_s(\Phi^r)$ is the layer sparsity loss and $\lambda^r$ is the Lagrangian multiplier.
This enables the network to decide what the \textit{depth} should be and which branch can be skipped. The combined loss for filter, layer and branch pruning can then be defined as:
\begin{equation}
\mathcal{L}(W,\Phi) = \mathcal{L}_D(W,\Phi) + \lambda^f\mathcal{L}_s(\Phi^f) + \lambda^l\mathcal{L}_s(\Phi^l) + \lambda^r\mathcal{L}_s(\Phi^r)
\end{equation}


\subsection{Block Pruning}
In our method the concept of block pruning is also very similar to that of layer and branch pruning. Let us define a multi-branched block as: %
\begin{equation}
\mathbf{b}_{l-1} = \mathbf{r}_{1,l-1}:\mathbf{r}_{k,l-1}
\end{equation}
In case of block pruning we also add shortcut connection between blocks if they are not already present (e.g : inception) and a learnable scaling factor at the end of each block such that the output of the entire block is multiplied with the branch pruning parameter $\phi^b$ and can result in eliminating the entire block. In such cases the input to the branch $\mathbf{h}_{l-1}$ can directly become input to the convolution layer immediately after the block as defined below
\begin{equation}
\mathbf{h}_l = f(\mathbf{b}_{l-1})g(\Phi_{l}^b,\epsilon))+\mathbf{h}_{l-n}
\end{equation}
We then extended the sparse dropout learning for block pruning by defining a block pruning loss function as:
\begin{equation}
\mathcal{L}(W,\Phi) = \mathcal{L}_D(W,\Phi) + \lambda^b\mathcal{L}_s(\Phi^b)
\end{equation}
where $\mathcal{L}_s(\Phi^b)$ is the layer sparsity loss and $\lambda^b$ is the Lagrangian multiplier.
This enables the network to decide which blocks can be pruned. The combined loss for filter, layer, branch and block pruning can then be defined as:
\begin{dmath}
\mathcal{L}(W,\Phi) = \mathcal{L}_D(W,\Phi) + \lambda^f\mathcal{L}_s(\Phi^f) + \lambda^l\mathcal{L}_s(\Phi^l) + \lambda^r\mathcal{L}_s(\Phi^r) + \lambda^b\mathcal{L}_s(\Phi^b).
\end{dmath}
This results in a comprehensive pruning of a network which is possible through our proposed approach. Furthermore, in our proposed method it is not necessary to use all the pruning strategies simultaneously, instead any of the $15$ combinations can be used such as filter and branch pruning or filter and layer and block pruning etc. For example on a sequential model one can implement filter pruning with one regularizer and layer pruning with a different regularizer so that the network is rewarded to prune both layers and filters from the network. In other words our approach provides $15$ different pruning strategies depending upon the network design and application related needs.

\section{Experimental Setup}
\subsection{Datasets}
We evaluated our approach on three standard benchmark datasets MNIST~\cite{13}, Fashion MNIST~\cite{xiao2017fashion} and CIFAR-10~\cite{12} . Both MNIST\cite{13} and Fashion MNIST\cite{xiao2017fashion} are small $10$ class datasets having $60,000$ training and $10,000$ test samples of size $28\times 28$. MNIST is a handwitten digits dataset, Fashion MNIST instead contains samples of apparels such as trouser, skirt, bag, ankle boot etc. The CIFAR-10 is another commonly used benchmarking dataset consisting of $6000$ and $600$ colour images (resolution $32\times 32$) per class respectively including $1000$ and $100$ training samples per class respectively. CIFAR-10 contains samples of 4 vehicle and 6 animal classes \footnote{http://groups.csail.mit.edu/vision/TinyImages/}.

\subsection{Models}
We tested our approach on VGG-16~\cite{15} and ResNet-56~\cite{15} models which are commonly used in neural network compression literature. In our experiments we used a custom VGG-16 model with three changes i) we used $0.05$ L2 Norm applied to every trainable layer to fight weight inflation problem caused by the network to mitigate the effect of shrinking scaling factors, ii) we added BatchNorm layer after each trainable layer to increase training speed and iii) we used a Dropout of 0.5 after 1st FC layer to prevent overfitting. ResNet-56 is chosen as it is a multi-branched network with ResBlocks on one branch and residual connection on the other branch.

\subsection{Baseline Accuracies}
Before experimentation we ran all the datasets on both models to achieve baseline accuracy.
In the case of VGG-16 we normalized our datasets and trained the base network for 180 epochs, We used 128 as batch size, we chose the best accuracy of these epochs as our baseline accuracy. We achieved 89.27\% , 98.10\% and 92.75\% for CIFAR-10, MNIST and Fashion MNIST respectively.

For ResNet-56 we normalized data and trained the network for 250 epochs, We used batch size of 32, We achived 92.75\% 98.56\%  and 94.1\% for CIFAR-10, MNIST and Fashion MNIST respectively.

\section{Results and Evaluation}

\subsection{Filter Pruning}
We implemented our filter pruning strategy using scaling factors on VGG-16. To do this we placed a scaling factor pruning tensor vectors of shape ($1\times$ num of filters) and range (0,1) in front of all convolution layers. We then applied L1 regularization with (weight $= 1$/num of filters) individually to all pruning tensor vectors and trained the network for $100$ epochs to estimate the parameter for each pruning factor during training. After training the network we empirically find the desired threshold for pruning filters and then pruned all the filters below that threshold using Keras-Surgeon\footnote{https://github.com/BenWhetton/keras-surgeon}, we also remove all the pruning tensors from the network. After obtaining this pruned network we fine-tune it for $10$ epochs to re-achieve accuracy. 

The results of our filter pruning strategy on all three datasets are shown in Table~\ref{tab:filterpruning}. Our results were comparable to the state-of-the-art results, GAN pruning\cite{lin2019towards} which achieved $82.2\%$ compression vs our 83.05\%. We noticed that filters from all over the network were pruned, however we observed uneven distribution of filter pruning where more filters were pruned from later layers as compared to earlier layers. For e.g. from in the CIFAR-10 pruned network only 6 and 2 filters were eliminated from first and second convolution respectively whereas from last two layers 462 and 398 filters were pruned. There are two reasons for this behavior i) VGG-16 has higher number of filters in later layers as compared to earlier layers and ii) earlier layers capture low-level features which has wide-scale usage across the network in extracting high-level features whereas not all the high-level features are useful for the task at hand which in this case is classification.
\begin{table}[t]
\caption{Filter Pruning on VGG-16}
\centering
{%
\begin{tabular}{lccc}
\hline
Dataset & Accuracy &  Params$\downarrow$ & $\%$ Flops$\downarrow$ \\ \hline
{CIFAR-10} & {88.27$\%$} & {83.05$\%$} & {40.21$\%$} \\
{MNIST} & {97.57$\%$} & {97.36$\%$} & {89.60$\%$} \\
{Fashion MNIST} & {92.16$\%$} & {92.95$\%$} & {69.16$\%$} \\
\hline
\end{tabular}}
\label{tab:filterpruning}
\end{table}

\subsection{Layer Pruning}
To examine our premise that pruning factors can prune the depth of the network too we created a custom VGG-16 network which had residual connections across all layers except at 4 cases of dimension mismatch. Although dimension mismatch can be handled via pooling and convolution by $1\times 1 \times d_l$ or by using mixed connections~\cite{27}, however for simplicity we chose not to handle the dimension mismatch. In addition to this the nature of our original network inherently had just 4 cases of dimension mismatch resulting in significant room in the network to prune. After creating these residual connections we added a pruning factor in front of all the layers with residual connections such that the output of the current layer is multiplied to a pruning factor before being added to the residual connection. We then apply Regularization with weightage (1/total num of prunable) to make these pruning factors sparse. Zeroing out these pruning factors means removing that whole layer from the network.

When we have our prunable model with additional residual connections and pruning factors, we train this network for $100$ epochs from scratch and estimate the parameter for each pruning factor during training. After this step we empirically find the pruning threshold and then remove layers from the network along with all the pruning factors. We then fine tune this network for $10$ epochs to re-achieve desired accuracy.

The results of our layer pruning strategy on CIFAR-10, MNIST and Fashion\_MNIST are shown in Table~\ref{tab:layerpruning}.

\begin{table}[t]
\caption{Table showing results of layer pruning on VGG-16}
\centering
\begin{tabular}{lccc}
\hline
Dataset & Accuracy & $\%$ Params$\downarrow$ & $\%$Flops$\downarrow$ \\ \hline
{CIFAR-10} & {87.77$\%$}  &  $52.16\%$ & $33.1\%$ \\
{MNIST} & {$98.82\%$} & {87.62$\%$} & {69.47$\%$} \\
{Fashion MNIST} & {92.80$\%$} & {87.62$\%$} & {69.47$\%$} \\ \hline
\end{tabular}
\label{tab:layerpruning}
\end{table}

\subsection{Branch Pruning}
In this work we also show that pruning of entire branch can be pruned from a neural network by using scaling factors. For this experiment we chose ResNet-56 because it is a multi-branched network, on one branch whole ResBlock is present while the other branch contains residual connection from previous block, both the branches merge after each block. We placed scaling factors on ResBlock branches to prune whole ResBlock from the network.
We then trained the network for $250$ epochs on CIFAR-10 dataset on which it reached over $91\%$ accuracy while the accuracy of unpruned network is reported to be around $92 \%$. After training we empirically found the optimal threshold to prune the network without losing accuracy. Our method drooped $17$ out of $27$ ResBlocks. The results of branch pruning are shown in Table~\ref{tab:BlockvsSOA} under the name \textbf{Our B}.

\subsection{Simultaneous Pruning of Filters and Layers}
In these experiments we applied scaling factors at two different granularities, one on the filter level and other on the layer level. To do this we used the same custom VGG-16 with additional residual connections that we used in the Layer pruning experiment. We added filter pruning scaling tensor vectors in front of all the conv layers similar to ones used in the filter pruning experiment. Two differently weighted regularizations were applied to the network (see Eq.~\ref{eq:filterlayer}), one for filter pruning scaling factor tensor vectors and one for layer pruning scaling factors. We then trained this custom VGG-16 for $100$ epochs.

After training we first searched for optimal thresholds for both filters and layers, we then changed the values below threshold to zero and above threshold to one. We then fine tuned this network for 10-20 epochs. We did this because in this experiment we removed all the pruning factors from the networks so pruning factors should be either 0 or 1 for the final pruning step, any value between (0,1) would mean that scaling factor was providing scaling to corresponding structure(e.g. filter, layers) but if we change these value to either 0 which will mean removal of corresponding structure or 1 which will presence of the corresponding structure without providing any scaling

After training we do the final pruning step where we first prune layers, to do this we use the corresponding scaling factor to either prune the layer or the residual connection, we remove the residual connections too because we do not want to have any foreign layer in the pruned network from the original network. After pruning layers we prune filters. We then fine tune the final network for $20$ epochs, for this experiment we need higher fine tuning because we removed the residual connections which changes the network graph.

The results of combined filter and layer pruning are shown in Table~\ref{tab:filterlayer}. In all of the experiments we re-achieved baseline accuracy within $1\%$ of margin.
\begin{table}[t]
\caption{Table showing results of simultaneous pruning of filters and layers on VGG-16}
\centering
\begin{tabular}{lccc}
\hline
Dataset & Accuracy & $\%$ Params$\downarrow$ & $\%$Flops$\downarrow$ \\ \hline
{CIFAR-10} & {88.11$\%$} & {89.66$\%$} & {57.79$\%$} \\
{MNIST} & {97.48$\%$} & {98.47$\%$} & {89.95$\%$} \\
{Fashion MNIST} & {92.10$\%$} & {97.99$\%$} & {83.44$\%$} \\ \hline
\end{tabular}
\label{tab:filterlayer}
\end{table}


\begin{table}[t]
\centering
\caption{Comparison of Filter Pruning and Filter and Layer Pruning with state-of-the-art methods on CIFAR-10 dataset using VGG-16 architecture. The table shows the change in overall number of flops, parameters and accuracy.}
\centering
\begin{tabular}{lllc}
\hline
Pruning & Accuracy &  Params$\downarrow$ & $\%$ Flops$\downarrow$ \\\hline
GAN Pruning~\cite{GANPruningCVPR2019} &$\approx 90\%$ & 82.2$\%$ & 45.20$\%$\\ 
Filter &88.27$\%$ & 83.05$\%$ & 40.21$\%$\\
Filter+Layer &88.11$\%$& \textbf{89.66$\%$} & 57.79$\%$\\ \hline
\end{tabular}
\label{tab:SOA2}
\end{table}

\begin{table*}[t]
\caption{Comparison with state-of-the-art methods on CIFAR-10 dataset using ResNet-56 architecture. The table shows the change in overall number of flops, parameters and accuracy.}
\centering
\resizebox{1.0\textwidth}{!}
{%
\begin{tabular}{lcccccccccc}
\hline
Metric & Li et al. \cite{25HaoICLR2017} & NISP \cite{52RuichiCVPR2018} & DCP-A \cite{56ZhuangweiNeurIPS2018} & CP \cite{15YihuiICCV2017} & AMC \cite{14YihuiECCV2018} & C-SGD \cite{6XiaohanArXiv2019} & GBN-40\cite{GateNIPS2019_8486} & GBN-30\cite{GateNIPS2019_8486} & Our B & Our B+F \\ \hline
FLOPS $\downarrow$ $\%$ & 27.6 & 43.6 & 47.1 & 50.0 & 50.0 & 60.8 & 60.1 & 70.3 & 63.78 & \textbf{79.00} \\
Params $\downarrow$ $\%$ & 13.7 & 42.6 & 70.3 & - & - & - & 53.5 & 66.7 & 59.52 & \textbf{78.14} \\
Accuracy & {-0.02} & {0.03} & \textbf{-0.01} & {1} & {0.90} & {-0.23} & {-0.33} & {0.03} & {0.97} & 1.35 \\ \hline
\end{tabular}
}
\label{tab:BlockvsSOA}
\end{table*}

\begin{table}[t]
\caption{Table showing results of simultaneous pruning of layers and branches ResNet-56}
\centering
\begin{tabular}{lccc}
\hline
Dataset & Accuracy &  Params$\downarrow$ & $\%$ Flops$\downarrow$ \\ \hline
CIFAR-10 & $92.13\%$ & $69.52\%$ & $66.50\%$ \\
{MNIST} & {98.28$\%$} & {81.19$\%$} & {84.60$\%$} \\
{Fashion MNIST} & {93.73$\%$} & {74.15$\%$} & {78.02$\%$} \\ \hline
\end{tabular}
\label{tab:layerblk}
\end{table}

\begin{table}[b]
\caption{Table showing results of simultaneous pruning of filters and branches on ResNet-56}
\centering
\begin{tabular}{lccc}
\hline
Dataset & Accuracy & $\%$ Params$\downarrow$ & $\%$Flops$\downarrow$ \\ \hline
{CIFAR-10} & {91.75$\%$} & $78.14\%$  &  $79.00\%$ \\
\end{tabular}
\label{tab:filterblock}
\end{table}

\subsection{Simultaneous Pruning of Layers and Branches}

For this experiment we placed two pruning factors in the ResBlock, one in front of the first layer and the other in front of the whole ResBlock. We then trained this network for $250$ epochs. After training we produced the pruned network in a similar fashion as done for layer block pruning. In this experiment we didn't remove the remaining scaling factors from the network so no fine tuning was required.Baseline accuracies are the same as stated . Results on all three datasets for combined layer and block pruning is shown in Table~\ref{tab:layerblk}.

\subsection{Simultaneous Pruning of Filters and Branch}
Similar to filter and layer pruning we also attempted to simultaneously prune filters and branches. In this experiment we used branch pruning version of ResNet-56 and added filter pruning to the first layer of each ResBlock. We then trained this network for $250$ epochs. After training we produced the pruned network in a similar fashion as done for layer branch pruning. We then froze all the layers in the network except BatchNorm and Dense and fine tuned it for $30$ epochs. We then remove filters from the network based on pruning tensor values.

Results on the CIFAR-10 dataset are shown in Table~\ref{tab:filterblock}. Baseline accuracy of unpruned network on CIFAR-10 dataset was $91\%$ to $92\%$ and the accuracy of our final pruned network was within $1.5\%$ margin of the baseline.

\subsection{Comparison with State-of-the-art}
We compared our proposed approach with nine different state of the art methods namely Li et al. \cite{25HaoICLR2017}, NISP~\cite{52RuichiCVPR2018}, DCP-A~\cite{56ZhuangweiNeurIPS2018}, CP~\cite{15YihuiICCV2017}, AMC~\cite{14YihuiECCV2018}, C-SGD~\cite{6XiaohanArXiv2019}, GBN-40~\cite{GateNIPS2019_8486}, GBN-30~\cite{GateNIPS2019_8486} and GAN Pruning~\cite{GANPruningCVPR2019}. Table~\ref{tab:SOA2} shows the comparison with these methods on CIFAR-10 datasets on VGG-16 architecture. It shows the comparison of our filter pruning and filter and layer pruning techniques vs. generative adversarial network~\cite{GANPruningCVPR2019}. Both our techniques performed better than SOA without noticeable loss in accuracy. Our filter and layer pruning resulted in an increase of $7.46\%$ in  parameter compression ratio and $12.59\%$ increase in flop compression ratio. 

Table~\ref{tab:BlockvsSOA} shows the comparison with these methods on CIFAR-10 datasets on ResNet-56 architecture. It shows the comparison of our block pruning and filter and block pruning techniques. It can be seen that one of our methods is second highest in terms of number of flops, third highest in terms of number of parameter reduction but our second method  which is a combination of filter and layer pruning, It can be seen that both our models performed worst in maintaining accuracy, this is partly due to difference in baseline accuracy between our experimentation and SOA table \cite{GateNIPS2019_8486}, Our experimentation achieved $92.75\%$ accuracy but SOA~\cite{GateNIPS2019_8486} achieved $93.1\%$ and these results are adjusted to SOA baseline accuracy but our method performed better then SOA in parameters and flops compression. We believe that this drop in accuracy can be reduced by further fine tuning the perimeters.


\section{Conclusions and Future Work} 
In this work we presented an idea of comprehensive pruning of neural networks which leverages from the idea of dropout and residual shortcut connections. We propose pruning at 4 different granularities i.e. filter, layers, branch (sequence of layers) and blocks (multi-branch blocks such as ResBlock, Inception-ResNet~\cite{szegedy2016rethinking} or those found in multi-task learning). Our pruning strategy is based on introducing learnable scaling factors at these granularities, these parameters are learned in an online manner during training and helps in generating a pruned network. In our case instead of using filter pruning as a proxy for layer pruning, we instead model pruning at different granularities by introducing separate sparsity loss for each of the granularity. 

We showed results on pruning at different granularities as well as their multiple combinations. We showed that by explicitly modeling pruning at these granularities, the network tries to increase the level of pruning without significant loss in accuracy. In future we aim to extend our scope of analysis to bigger datasets such as CIFAR-100 and on more complicated networks such as Inception-ResNets~\cite{szegedy2016rethinking}. We will also be analysing the effect of using different schemes of regularizations to identify the best at each granularity.

\newpage
{\small
\bibliographystyle{ieee_fullname}

\begin{thebibliography}{9}
\bibitem{1}
Chechik, Gal, Isaac Meilijson, and Eytan Ruppin. "Synaptic pruning in development: a computational account." Neural Computation: 1759-1777, 1998

\bibitem{2}
LeCun, Yann, John S. Denker, and Sara A. Solla. "Optimal brain damage." Advances in neural information processing systems, 1990.

\bibitem{3}
Han, Song, et al. "Learning both weights and connections for efficient neural network." Advances in neural information processing systems, 2015.

\bibitem{4}
Li, Hao, et al. "Pruning filters for efficient convnets." arXiv preprint arXiv:1608.08710, 2016.

\bibitem{5}
Sun, Xinlu, et al. "Pruning filters with L1-norm and standard deviation for CNN compression." Eleventh International Conference on Machine Vision (ICMV 2018). Vol. 11041. International Society for Optics and Photonics, 2019.

\bibitem{6}
Singh, Arshdeep, Padmanabhan Rajan, and Arnav Bhavsar. "Deep hidden analysis: A statistical framework to prune feature maps." ICASSP 2019-2019 IEEE International Conference on Acoustics, Speech and Signal Processing. IEEE, 2019.

\bibitem{7}
He, Yang, et al. "Filter pruning via geometric median for deep convolutional neural networks acceleration." Proceedings of the IEEE Conference on Computer Vision and Pattern Recognition, 2019.

\bibitem{8}
Hu, H., et al. "A data-driven neuron pruning approach towards efficient deep architectures." arXiv preprint arXiv:1607.03250, 2016.

\bibitem{9}
Ding, Xiaohan, et al. "Centripetal sgd for pruning very deep convolutional networks with complicated structure." Proceedings of the IEEE Conference on Computer Vision and Pattern Recognition, 2019.

\bibitem{10}
Dubey, Abhimanyu, Moitreya Chatterjee, and Narendra Ahuja. "Coreset-based neural network compression." Proceedings of the European Conference on Computer Vision, 2018.

\bibitem{11}
He, Kaiming, et al. "Deep residual learning for image recognition." Proceedings of the IEEE conference on computer vision and pattern recognition, 2016.

\bibitem{12}
Dataset collected by Alex Krizhevsky, Vinod Nair, and Geoffrey Hinton. https://www.cs.toronto.edu/~kriz/cifar.html

\bibitem{13}
Yann LeCun, Courant Institute, NYU, Corinna Cortes, Google Labs, New York, Christopher J.C. Burges, Microsoft Research, Redmond. MNIST: http://yann.lecun.com/exdb/mnist/

\bibitem{xiao2017fashion}
  Han Xiao, Kashif Rasul and Roland Vollgraf. "Fashion-MNIST: a novel image dataset for benchmarking machine learning algorithms." arXiv preprint arXiv:1708.07747, 2017.
  
\bibitem{15}
Simonyan, Karen, and Andrew Zisserman. "Very deep convolutional networks for large-scale image recognition." arXiv preprint arXiv:1409.1556, 2014.

\bibitem{16}
Chin, T. W., Ding, R., Zhang, C., and Marculescu, D. "Legr: Filter pruning via learned global ranking." arXiv preprint arXiv:1904.12368, 2019.

\bibitem{17}
Lin, M., Ji, R., Wang, Y., Zhang, Y., Zhang, B., Tian, Y., and Shao, L. "HRank: Filter Pruning using High-Rank Feature Map." arXiv preprint arXiv:2002.10179, 2020

\bibitem{18}
He, Yihui, Xiangyu Zhang, and Jian Sun. "Channel pruning for accelerating very deep neural networks." Proceedings of the IEEE International Conference on Computer Vision, 2017.

\bibitem{20}
Luo, Jian-Hao, and Jianxin Wu. "Neural Network Pruning with Residual-Connections and Limited-Data." arXiv preprint arXiv:1911.08114, 2019.

\bibitem{21}
Guo, Shaopeng, et al. "DMCP: Differentiable Markov Channel Pruning for Neural Networks." arXiv preprint arXiv:2005.03354, 2020.

\bibitem{22}
Ye, Jianbo, et al. "Rethinking the smaller-norm-less-informative assumption in channel pruning of convolution layers." arXiv preprint arXiv:1802.00124, 2018.

\bibitem{23}
Zhu, Michael, and Suyog Gupta. "To prune, or not to prune: exploring the efficacy of pruning for model compression." arXiv preprint arXiv:1710.01878, 2017.

\bibitem{24}
Zhuang, Zhuangwei, et al. "Discrimination-aware channel pruning for deep neural networks." Advances in Neural Information Processing Systems, 2018.

\bibitem{25}
Dong, Xuanyi, and Yi Yang. "Network pruning via transformable architecture search." Advances in Neural Information Processing Systems, 2019.

\bibitem{26}
Zhou, Yuefu, et al. "Accelerate CNN via Recursive Bayesian Pruning." Proceedings of the IEEE International Conference on Computer Vision, 2019.

\bibitem{27}
Lemaire, Carl, Andrew Achkar, and Pierre-Marc Jodoin. "Structured pruning of neural networks with budget-aware regularization." Proceedings of the IEEE Conference on Computer Vision and Pattern Recognition, 2019.

\bibitem{JMLR:v15:srivastava14a}
  Nitish Srivastava and Geoffrey Hinton and Alex Krizhevsky and Ilya Sutskever and Ruslan Salakhutdinov. "Dropout: A Simple Way to Prevent Neural Networks from Overfitting." Journal of Machine Learning Research, 2014.

\bibitem{NIPS2015_5666}
Kingma, Durk P and Salimans, Tim and Welling, Max. "Variational Dropout and the Local Reparameterization Trick." Advances in Neural Information Processing Systems, 2015.

\bibitem{yolo}
Redmon, Joseph, et al. "You only look once: Unified, real-time object detection." Proceedings of the IEEE conference on computer vision and pattern recognition, 2016.

\bibitem{tinyyolo}
Huang, Rachel, Jonathan Pedoeem, and Cuixian Chen. "YOLO-LITE: a real-time object detection algorithm optimized for non-GPU computers." 2018 IEEE International Conference on Big Data (Big Data). IEEE, 2018.

\bibitem{resnet}
He, Kaiming, et al. "Deep residual learning for image recognition." Proceedings of the IEEE Conference on Computer Vision and Pattern Recognition, 2016.

\bibitem{inception}
Szegedy, Christian, et al. "Inception-v4, Inception-ResNet and the impact of residual connections on learning." arXiv preprint arXiv:1602.07261 (2016).

\bibitem{alexnet}
Krizhevsky, Alex, Ilya Sutskever, and Geoffrey E. Hinton. "{ImageNet} classification with deep convolutional neural networks." Advances in neural information processing systems, 2012.

\bibitem{HintonPaper}
Hinton, Geoffrey, Oriol Vinyals, and Jeff Dean. "Distilling the knowledge in a neural network." arXiv preprint arXiv:1503.02531, 2015.

\bibitem{GANPruningCVPR2019}
Shaohui Lin, Rongrong Ji, Chenqian Yan, Baochang Zhang, Liujuan Cao, Qixiang Ye, Feiyue Huang, David S. Doermann. "Towards Optimal Structured CNN Pruning via Generative Adversarial Learning." Proceedings of the IEEE Conference on Computer Vision and Pattern Recognition, 2019.

\bibitem{LeCun1998}	
Y. LeCun, L. Bottou, Y. Bengio and P. Haffner. "Gradient-Based Learning Applied to Document Recognition." Proceedings of the IEEE, 86(11):2278-2324, November 1998.

\bibitem{wang2018adversarial}
  Yunhe Wang,  Chang Xu, Chao Xu and Dacheng Tao. "Adversarial learning of portable student networks. "Thirty-Second AAAI Conference on Artificial Intelligence, 2018.

\bibitem{hinton2015distilling}
Geoffrey Hinton, Oriol Vinyals and Jeff Dean. "Distilling the knowledge in a neural network." NIPS Deep Learning and Representation Learning Workshop, 2015.

\bibitem{you2017learning}
You, Shan and Xu, Chang and Xu, Chao and Tao, Dacheng. "Learning from multiple teacher networks." 23rd ACM SIGKDD International Conference on Knowledge Discovery and Data Mining, 2017.

\bibitem{heo2019comprehensive}
  Byeongho Heo, Jeesoo Kim, Sangdoo Yun, Park Hyojin,  Nojun Kwak and Jin Young Choi "A comprehensive overhaul of feature distillation." IEEE International Conference on Computer Vision, 2019.

\bibitem{NIPS2018_7980}
Xu Lan, Xiatian Zhu and Shaogang Gong. "Knowledge Distillation by On-the-Fly Native Ensemble." Advances in Neural Information Processing Systems, 2018.

\bibitem{lin2019towards}
Shaohui Lin, Rongrong Ji, Chenqian Yan, Baochang Zhang, Liujuan Cao, Qixiang Ye, Feiyue Huang and David Doermann "Towards Optimal Structured CNN Pruning via Generative Adversarial Learning." Proceedings of the IEEE Conference on Computer Vision and Pattern Recognition, 2019.

\bibitem{25HaoICLR2017}
Hao Li, Asim Kadav, Igor Durdanovic, Hanan Samet, and Hans Peter Graf. Pruning filters for efficient convnets. In International Conference on Learning Representations, 2017.

\bibitem{52RuichiCVPR2018}
 Ruichi Yu, Ang Li, Chun-Fu Chen, Jui-Hsin Lai, Vlad I. Morariu, Xintong Han, Mingfei Gao, ChingYung Lin, and Larry S. Davis. NISP: pruning networks using neuron importance score propagation. In Conference on Computer Vision and Pattern Recognition, pages 9194–9203, 2018

\bibitem{56ZhuangweiNeurIPS2018}
Zhuangwei Zhuang, Mingkui Tan, Bohan Zhuang, Jing Liu, Yong Guo, Qingyao Wu, Junzhou Huang, and
Jin-Hui Zhu. Discrimination-aware channel pruning for deep neural networks. In Advances in Neural Information Processing Systems, pages 883–894, 2018.

\bibitem{15YihuiICCV2017}
Yihui He, Xiangyu Zhang, and Jian Sun. Channel pruning for accelerating very deep neural networks. In IEEE International Conference on Computer Vision, pages 1398–1406, 2017.

\bibitem{14YihuiECCV2018}
Yihui He, Ji Lin, Zhijian Liu, Hanrui Wang, Li-Jia Li, and Song Han. AMC: {AutoML} for model compression and acceleration on mobile devices. In European Conference on Computer Vision, pages 815–832, 2018.

\bibitem{6XiaohanArXiv2019}
Xiaohan Ding, Guiguang Ding, Yuchen Guo, and Jungong Han. Centripetal SGD for pruning very deep
convolutional networks with complicated structure. arXiv, abs/1904.03837, 2019.

\bibitem{GateNIPS2019_8486}
Zhonghui You, Kun Yan, Jinmian Ye, Meng Ma and Ping Wang. "Gate Decorator: Global Filter Pruning Method for Accelerating Deep Convolutional Neural Networks." Advances in Neural Information Processing Systems, 2019.
  
\bibitem{szegedy2016rethinking}
	Christian Szegedy, Vincent Vanhoucke, Sergey Ioffe, Jon Shlens and Zbigniew Wojna. "Rethinking the inception architecture for computer vision." IEEE Conference on Computer Vision and Pattern Recognition, 2818--2826, 2016.
	
\end{thebibliography}

}

\end{document}